\def\year{2020}\relax
\documentclass[letterpaper]{article} 
\usepackage{aaai20}  
\usepackage{times}  
\usepackage{helvet} 
\usepackage{courier}  
\usepackage[hyphens]{url}  
\usepackage{graphicx} 
\usepackage{graphicx}  
\usepackage{color}
\usepackage{multirow}
\usepackage{multicol}
\usepackage{latexsym}
\usepackage{mathtools}

\definecolor{darkgreen}{rgb}{0.13, 0.55, 0.13}

\title{Adapting Language Models for Non-Parallel Author-Stylized Rewriting}
\author{Bakhtiyar Syed\textsuperscript{$\ddagger$}, Gaurav Verma\textsuperscript{$\dagger$}, Balaji Vasan Srinivasan\textsuperscript{$\dagger$} \\\Large{\textbf{Anandhavelu Natarajan\textsuperscript{$\dagger$}, Vasudeva Varma\textsuperscript{$\ddagger$}}}\\
{\textsuperscript{$\ddagger$}{IIIT Hyderabad}, \textsuperscript{$\dagger$}{Adobe Research}}\\
syed.b@research.iiit.ac.in\\
\{gaverma, balsrini, anandvn\}@adobe.com\\
vv@iiit.ac.in
}
 
\begin{document}

\maketitle

\begin{abstract}
Given the recent progress in language modeling using Transformer-based neural models and an active interest in generating stylized text, we present an approach to leverage the generalization capabilities of a language model to rewrite an input text in a target author's style. Our proposed approach adapts a pre-trained language model to generate author-stylized text by fine-tuning on the author-specific corpus using a denoising autoencoder (DAE) loss in a cascaded encoder-decoder framework. Optimizing over DAE loss allows our model to learn the nuances of an author's style \textit{without} relying on parallel data, which has been a severe limitation of the previous related works in this space. To evaluate the efficacy of our approach, we propose a linguistically-motivated framework to quantify stylistic alignment of the generated text to the target author at lexical, syntactic and surface levels. The evaluation framework is both interpretable as it leads to several insights about the model, and self-contained as it does not rely on external classifiers, e.g. sentiment or formality classifiers. Qualitative and quantitative assessment indicates that the proposed approach rewrites the input text with better alignment to the target style while preserving the original content better than state-of-the-art baselines.
\end{abstract}

\section{Introduction}
There has been a growing interest in studying style in natural language and solving tasks related to it  \cite{hu2017,shen2017style,sandeep2018multipleattribute,fu2018style,vadapalli-etal-2018-science,niu2018polite}. Tasks like genre classification \cite{kessler1997automatic}, author profiling \cite{garera2009modeling}, sentiment analysis \cite{wilson2005recognizing}, social relationship classification \cite{peterson2011email} have been of active interest to the community. Recently, stylized text generation \cite{hovy1990pragmatics,inkpen2006building} and style transfer \cite{li2018delete,prabhumoye2018style,fu2018style} have gained traction; both these tasks aim to generate realizations of an input text that align to a target style. A majority of the work here is focused around generating text with different levels of sentiment \cite{shen2017style,ficler2017controlling} and formality \cite{jain2019unsupervised} and also a combination of these attributes \cite{sandeep2018multipleattribute}. The interest along these lines has given rise to annotated and parallel data that comprise of paired realizations that lie on opposite ends of formality and sentiment spectrum \cite{rao2018dear,mathews2016senticap}. The dimensions of style considered across all these works are psycholinguistic aspects of text and the aim is to transfer the text across different levels of the chosen aspect. 

\begin{figure}[t]
\centering
\includegraphics[width=1.0\columnwidth]{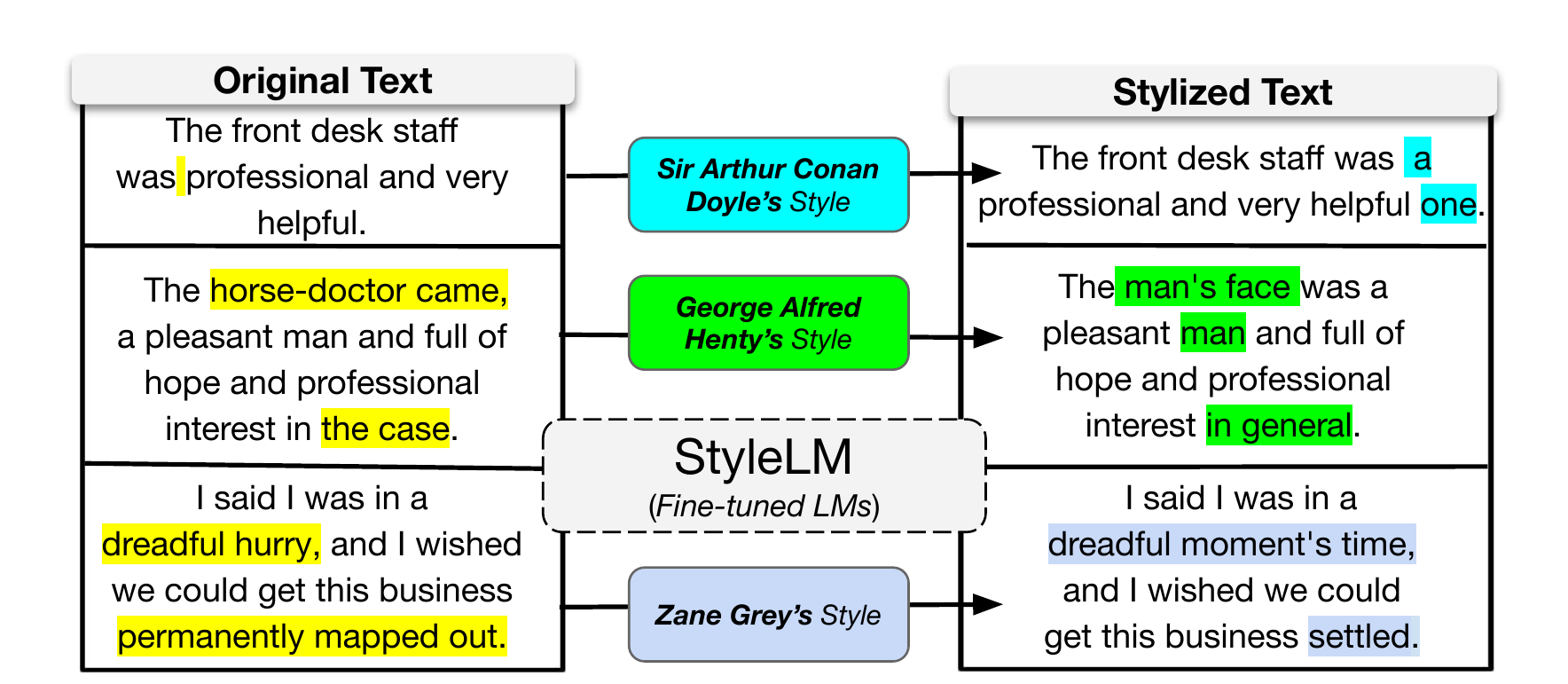}
\caption{\small{An overview of generating author-stylized text using \textit{StyleLM}, our proposed model.}}
\label{fig:teaser}
\end{figure}

However, there has been lack of explorations that aim to generate text across author styles -- wherein the notion of style is not a specific psycholinguistic aspect but an amalgam of the author's linguistic choices expressed in their writing \cite{jhamtani2017shakespearizing,tikhonov2018guess}. While the work by \citeauthor{jhamtani2017shakespearizing} (\citeyear{jhamtani2017shakespearizing}) tries to generate ``Shakespearized'' text from Modern English and is in a similar vein, it relies on the availability of parallel data. Since the availability of parallel data is not always guaranteed and it is arduous to curate one, such an approach cannot scale for different authors. We therefore propose a novel framework for author-stylized rewriting without relying on parallel data with source-text to target-text mappings. Figure \ref{fig:teaser} shows a few examples where an input text is rewritten in the style of a chosen author by our model.

Our approach for generating author-stylized text involves leveraging the generalization capabilities of state-of-the-art language models and adapting them to incorporate the stylistic characteristics of a target author without the need of parallel data. We first pre-train a language model on a combination of author corpus \cite{lahiri:2014:SRW} and Wikipedia data using the masked language modeling objective \cite{devlin2019bert}. Drawing inspiration from the unsupervised machine translation setup of \citeauthor{lample2019cross} (\citeyear{lample2019cross}), we cascade two copies of this pre-trained language model into an encoder-decoder framework, where the parameters of the encoder and decoder are initialized with the pre-trained language model. This cascaded framework is fine-tuned on a specific target author's corpus of text by reconstructing the original text from its noisy version and optimizing on a denoising autoencoder loss. 
The fine-tuned model thus adapts itself towards the style of the target author as we show via our experimental analysis.

Author-stylized rewriting of text takes a text, which may or may not have a distinctive style, and rewrites it in a style that can be attributed to a target author. 
Since the writing style of authors is determined by several linguistically active elements that are expressed at lexical, syntactic, and semantic levels, it is challenging to evaluate the stylistic alignment of rewritten text to target author's style. To this end, we propose a novel and interpretable framework that is linguistically motivated, to quantify the extent of stylistic alignment at multiple levels. 
As we elaborate upon in the later sections, our evaluation suggests that the proposed approach performs better than three relevant and competitive baselines -- showing significant adaption to the writing style of target authors, both qualitatively and quantitatively. Notably, our approach performs on par (and better in certain dimensions) with state-of-the-art method for stylistic rewriting \textit{using parallel data}, \textit{without} leveraging the parallel nature of underlying data. 


The \textbf{key contributions of this work} are threefold.
\begin{enumerate}
  \item We propose and evaluate an approach to generate author-stylized text without relying on parallel data by adapting state-of-the-art language models.
  \item We propose an evaluation framework to assess the efficacy of stylized text generation that accounts for alignment of lexical and syntactic aspects of style. Contrary to existing evaluation techniques, our evaluation framework is linguistically-aware and easily interpretable.
  \item Our proposed approach shows significant improvement in author-stylized text generation over baselines, both in quantitative and qualitative evaluations.
\end{enumerate}

\section{Related Work}
\subsubsection{Stylized Text Generation:}
In recent times, several explorations that aim to generate stylized text define a psycholinguistic aspect, like, formality or sentiment \cite{shen2017style,ficler2017controlling,jain2019unsupervised} and transfer text along this dimension. The approaches themselves can range from completely supervised, which is contingent on the availability of parallel data \cite{ficler2017controlling}, to unsupervised \cite{shen2017style,li2018delete,jain2019unsupervised}. Some of the influential unsupervised approaches include \textit{(a)} using readily available classification-based discriminators to guide the process of generation \cite{fu2018style}, \textit{(b)} using simple linguistic rules to achieve alignment with the target style \cite{li2018delete}, or \textit{(c)} using auxiliary modules (called \textit{scorers}) that score the generation process on aspects like fluency, formality and semantic relatedness while deciding on the learning scheme of the encoder-decoder network \cite{jain2019unsupervised}. However, in the context of our setting, it is not possible to build a classification-based discriminator or scorers to generate author-stylized text. Moreover, linguistic-rule based generations are intractable given the large number of rules required to define a target author's style. To this end, we aim to adapt state-of-the-art language models to generate author-stylized text from non-parallel data. The choice of using language models is motivated by the fact that stylistic \textit{re}writing \textit{builds on} the task of simple text generation (i.e., writing).

There are some works that adapt an input text to the writing style of a specific author \cite{jhamtani2017shakespearizing,tikhonov2018guess}. While \citeauthor{tikhonov2018guess} (\citeyear{tikhonov2018guess}) generate author-stylized poetry by learning the style end-to-end using conditioning and concatenated embeddings of corresponding stylistic variables, theirs is not a \textit{rewriting} task. \citeauthor{jhamtani2017shakespearizing} (\citeyear{jhamtani2017shakespearizing}) aim to generate “Shakespearized” version of modern English language using parallel data. 
Our proposed approach aims to overcome this shortcoming by only relying on non-parallel data and only requires the corpus of the target author text for stylistic rewriting. As we show later, the proposed framework is comparable (even better in some of the dimensions) to \citeauthor{jhamtani2017shakespearizing}'s approach across content preservation and style transmission metrics without utilizing the parallel corpus.

\subsubsection{Language Models:}
Generative pre-training of sentence encoders \cite{radford2018improving,devlin2019bert,howard2018universal} has led to strong improvements on several natural language tasks. Their approach is based on learning a Transformer \cite{vaswani2017attention} language model on a large unsupervised text corpus and then fine-tuning on classification and inference-based natural language understanding (NLU) tasks. Building up on this, \citeauthor{lample2019cross} (\citeyear{lample2019cross}) extend this approach to learn cross-lingual language models. Taking inspiration from this, we extend the generative pre-training for our task of author-stylized rewriting. 

The recently proposed language model GPT-2 \cite{radford2019language} is pre-trained on a large and diverse dataset (WebText) and is shown to perform well across several domains and datasets including natural language generation. 
The unsupervised pre-training is setup to model the generation probability of the next word, given the previous words, i.e., $P(y_t \mid y_{1:t-1}, \mathbf{x})$ -- more generally referred to as the causal language modeling (CLM) objective. Specific to the task of text generation, it takes an input prompt ($\mathbf{x}$) and aims to generate text that adheres to the input context. As substantiated in the later sections, GPT-2, when fine-tuned on author-specific corpus, shows significant stylistic alignment with the writing style of target author. However, given the inherent differences involved in the setup of stylistic \textit{rewriting} and stylized text generation, it performs poorly on content preservation. While in stylistic rewriting, the objective is to retain the information in the input text in the stylized generation, stylistic generation by GPT-2 generates the content that is \textit{related} to the input \textit{prompt} and hence fine-tuned GPT-2 cannot address the task of stylistic rewriting. 
In the cross-lingual language modeling literature, a recent exploration by \citeauthor{lample2019cross} (\citeyear{lample2019cross}) learns cross-lingual language models by first pre-training on $3$ different language modelling objectives: \textit{(i)} causal language model (CLM), \textit{(ii)} masked language model (MLM) -- similar to BERT \cite{devlin2019bert}, and \textit{(iii)} translation language model (TLM) - which is a supervised setup leveraging parallel corpora. Following the pre-training, \citeauthor{lample2019cross} cascade the encoder and decoder 
 to address the tasks of \emph{supervised} cross-lingual classification and machine translation by fine-tuning on a combination of denoising auto-encoder (DAE) and back-translation losses. 
Taking inspiration from this work, we pre-train a language model on a large corpus using MLM objective and then fine-tune it on author-specific corpus using DAE loss in an encoder-decoder setup. Using DAE loss ensures that we don't rely on availability of parallel corpora, while the pre-trained language model facilitates the task of rewriting by building a firm substratum. 

\subsubsection{Evaluating Stylized Generation:}
\citeauthor{fu2018style} (\citeyear{fu2018style}) propose an evaluation framework to assess the efficacy of style transfer models on two axes: \textit{(i)} \textit{content preservation} and \textit{(ii)} \textit{transfer strength}. While the former caters to the content overlap between input and generated text (quantified using BLEU \cite{papineni2002bleu}), the latter takes into account the alignment of generated text with target style. In their setup, as it is with many others, the notion of target style is a psycholinguistic aspect (formality or sentiment) for which classifiers or scorers are readily available and are hence used to quantify the transfer strength \cite{jain2019unsupervised,li2018delete,mir2019evaluating}. However, for evaluating  \textit{author}-stylized text generations the evaluation frameworks are not well established. \citeauthor{jhamtani2017shakespearizing} (\citeyear{jhamtani2017shakespearizing}) and \citeauthor{tikhonov2018guess} (\citeyear{tikhonov2018guess}) overcome this by using the content preservation metrics as a proxy of transfer strength, leveraging the availability of the ground-truth stylized text. The unavailability of a suitable metric for transfer strength is particularly pronounced in evaluating unsupervised approaches as there is no target data to compare the generations against. To this end, we propose a linguistically-aware and interpretable evaluation framework which quantifies alignment of multiple lexical and syntactic aspects of style in the generated text with respect to the target author's style. 

\section{Proposed Approach: StyleLM}
There are two key aspects to our approach -- pre-training a Transformer-based language model on a large dataset that acts as a substratum and fine-tuning on author-specific corpus using DAE loss to enable stylized rewriting. The entire approach is \textbf{not} contingent on the availability of parallel data and the models are learned in a self-supervised manner.

\begin{figure*}[t]
    \centering
    \includegraphics[width=0.78\textwidth]{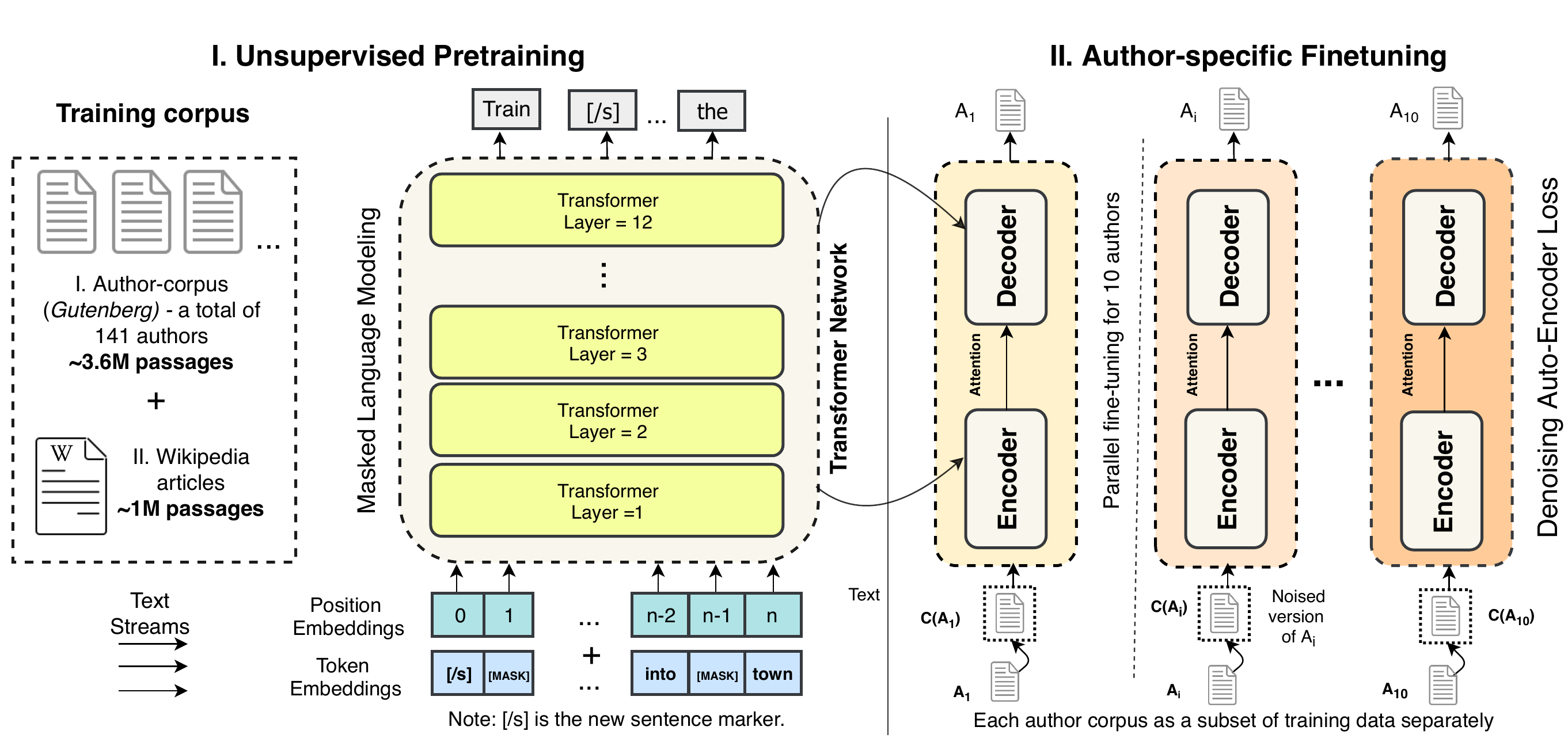}
    \caption{\small{\textbf{Proposed \textit{StyleLM} model}. We first pre-train a language model on large English corpus (\textit{I. Unsupervised Pretraining}) and then cascade the pre-trained LMs into an encoder-decoder like framework (as represented by the curved arrows). The encoder-decoder is fine-tuned separately on each of the target author's corpus using DAE loss  (\textit{II. Author-specific fine-tuning}). }}
    \label{fig:model}
\end{figure*}

Figure \ref{fig:model} illustrates the proposed framework for stylistic rewriting. We first pre-train the Transformer-based language model on a large unsupervised corpus using the masked language modeling (MLM) objective \cite{devlin2019bert}.
The choice of using a Transformer-based architecture is based on their recent success in language modeling \cite{vaswani2017attention,devlin2019bert,radford2018improving,radford2019language}. The MLM objective encourages the LM to predict the masked word(s) from the input sequence of words leveraging bidirectional context information of the input.  

Given a source sentence $\mathbf{x}$ , $\mathbf{x}^{\setminus u}$ is a modified version of $\mathbf{x}$ where its token from position $u$ is masked by replacing it with a mask token $[\mathbf{MASK}]$ - thus keeping the length of the masked sentence unchanged.
The MLM objective pre-trains the language model by predicting the original token $x^{u}$, taking the masked sequence $x^{\setminus u}$ as input, while learning the parameters $\theta$ for the conditional probability of the language model. We minimize the log-likelihood given by,\begin{equation}
	\begin{aligned}
	\small
	\label{equ_bt}
	L(\theta; \mathcal{X}) 
	& = \frac{1}{|\mathcal{X}|}\Sigma_{\mathbf{x} \in \mathcal{X}}\log P(\mathbf{x}^{u} \mid \mathbf{x}^{\setminus u};\theta)
	\end{aligned}\footnote{The equation given here describes MLM for one token \cite{song2019mass}. In practice, multiple tokens are masked in the original BERT architecture and for our experiments, which is just an extension of the above idea for training speed-up. }
\end{equation}
where, $\mathcal{X}$ denotes the entire training corpus. 
For pre-training the language model using the MLM objective, following \citeauthor{devlin2019bert} (\citeyear{devlin2019bert}), we randomly mask $15\%$ of the tokens in each input sequence,  replace them with the $[MASK]$ token $80\%$ of the time, by a random token $10\%$ of the time, and keep them unchanged $10\%$ of the time. A difference between our model and the MLM proposed by \citeauthor{devlin2019bert} (\citeyear{devlin2019bert}) is the use of text streams of sentences (truncated at $256$ tokens) in contrast to pairs of sentences. This has been shown to give considerable gains for text generation tasks \cite{lample2019cross}. Also, unlike \citeauthor{devlin2019bert} (\citeyear{devlin2019bert}), we do not use the \textit{Next Sentence Prediction} (NSP) objective.

The language model (LM) above learns to predict the masked words over a large corpus, but does not incorporate any style-related fine-tuning that facilitates rewriting the input text in a given target author's style. To achieve this, we cascade two instances of the pre-trained LM in an encoder-decoder setup where one instance acts as the encoder and the other acts as a decoder. In other words, the learnable parameters of both encoder and decoder are initialized using the pre-trained LM. Note that the architecture of Transformer-based language models allows two exact instances of the pre-trained LM to be cascaded, without explicitly aligning the encoder's output and the decoder's input \cite{bahdanau2014neural} since the attention-mechanism is inherent in the design of Transformers \cite{vaswani2017attention}.
\citeauthor{lample2019cross} (\citeyear{lample2019cross}) successfully used such a cascading to bootstrap the iterative process of the model initialization for the unsupervised machine translation task. Taking inspiration from this, we fine-tune the encoder-decoder on the DAE loss, given by,
\begin{eqnarray}
    \mathcal{L}^{DAE} & = & \mathbf{E}_{\mathbf{x} \sim \mathcal{S}} [-\log P(\mathbf{x} \mid C(\mathbf{x}))] 
\label{eq:DAEloss}
\end{eqnarray}
where, $C(x)$ is the noisy version of the input sentence $\mathbf{x}$ and $\mathcal{S}$ are the sentences in target author's corpus. To obtain a noisy version $C({\mathbf{x}})$ of input text $\mathbf{x}$, we drop every word in $\mathbf{x}$ with a probability $p_{drop}$ and also blank the input words with a probability $p_{blank}$\footnote{replace the word with $[BLANK]$.}. 

When the pre-trained language model is cascaded as the encoder and decoder, and further fine-tuned with a noisy version of the text, the encoder generates the masked words (since that is the original objective of the pre-trained LM). However, since the input to the decoder, which is same as the output of the encoder, has no masked words, it tries to reconstruct the clean version of the noisy input text. In other words, fine-tuning the encoder-decoder on target author's corpus using the DAE loss (equation \ref{eq:DAEloss}) pushes the model's decoder towards inducing target author's style while rewriting the input text from the encoder.


\subsubsection{Implementation Details}

During pre-training with MLM, we use the Transformer encoder \cite{vaswani2017attention} ($12$-layer) with GELU activations \cite{Hendrycks2017BridgingNA}, $512$ hidden units, $16$ heads, a dropout rate of $0.1$ and learned positional embeddings. We train our models with the Adam optimizer \cite{Kingma2014AdamAM}, and a learning rate  of $10^{-4}$ . We use streams of $256$ tokens and a mini-batches of size $32$. 
We train our model on the MLM objective until the language model's perplexity shows no improvement over the validation dataset.
For fine-tuning on a target author, which involves reconstruction of the \textit{whole} input passage\footnote{Unlike the MLM which predicts only a part of the input.} from its noisy version we use the same pre-trained MLM Transformer initialization for both the encoder and decoder, similar to \citeauthor{lample2019cross} (\citeyear{lample2019cross}), with the same hyperparameters used for pre-training. $p_{drop}$ and $p_{blank}$ are set to $0.1$ and the model is fine-tuned until convergence. 

To handle the vocabulary size for such a huge dataset, we use Byte Pair Encoding (BPE) \cite{Sennrich2015NeuralMT} on the combined training dataset and learn $80k$ BPE codes on the dataset. Since we use BPE codes on the combination of the training dataset of the $141$ authors, we can scale these for any author at will -- thus the ability to adapt to any author in the Gutenberg corpus or beyond.

\section{Evaluation Framework}
\subsubsection{Dataset} We collated a subset of the Gutenberg corpus \cite{lahiri:2014:SRW} consisting of $142$ authors and $2,857$ books written by them. For evaluating on a completely unseen author (a zero-shot setting), we set aside the writings by Mark Twain from the training corpus. The remaining authors are used as training corpus during pre-training resulting in $\sim 3.6$M passages. To diversify the pre-training dataset, we use $1$ million passages from Wikipedia \cite{radford2018improving} along with $\sim 3.6$M passages from the Gutenberg corpus -- leading to a total of $\sim4.6$M passages for pre-training the LM. Of these, we set aside $5000$ passages for validation and $5000$ for test during the pre-training stage. 

To fine-tune the encoder-decoder framework from the pre-trained LM, we pick a subset of $10$ authors from the Gutenberg corpus and independently treat them as target authors to generate author-stylized text. The $10$ chosen authors are: Sir Arthur Conan Doyle, Charles Dickens, George Alfred Henty, Nathaniel Hawthorne, Robert Louis Stevenson, Rudyard Kipling, Thomas Hardy, William Makepeace Thackeray, and Zane Grey. We fine-tune independently for each of the $10$ target authors and evaluate the efficacy of our proposed approach using a novel evaluation framework with roots in linguistic literature, described in a later section.

For inference during test-time, we use the following three corpora to obtain our source sentences : (a) texts from books written by Mark Twain, (b) Opinosis Review dataset \cite{ganesan2010opinosis}, (c) a Wikipedia article on \textit{Artificial Intelligence} (\url{https://en.wikipedia.org/wiki/Artificial_intelligence}) which does not appear in the original mix of the Wikipedia training corpus. Texts from these sources span a diverse range of topics and writing styles -- while Mark Twain's writings are literary, Opinosis reviews are everyday, the Wikipedia article on AI presents an interesting scenario where many of the words in the source text are not present in target author's corpus, given the different timelines.

We evaluate our performance against $4$ baselines - $3$ of which are trained on non-parallel data, while the $4$th one uses parallel data. 

\noindent\textbf{1. Vanilla GPT-2 based generation}:
\citeauthor{radford2019language} (\citeyear{radford2019language}) show that language models present considerable promise as unsupervised multi-task learners. 
We use their vanilla GPT-2 pre-trained Transformer decoder \cite{radford2019language} as our first baseline.\footnote{In our experimental setup, we utilise the pre-trained $124M$ parameter model for generation - \url{https://github.com/openai/gpt-2}}. The GPT-2 is fed a prompt directly during inference and the generated outputs are compared against other generations.

\noindent\textbf{2. Author fine-tuned GPT-2}:
The second baseline is the fine-tuned GPT-2 model for the cross-entropy loss on each of the target author's corpus separately. 
We use the stylized text generated by providing a prompt to the fine-tuned model for comparisons. 

\noindent\textbf{3. Denoising-LM : no author-specific fine-tuning}:
This baseline is similar to our StyleLM network, but fine-tuned on the entire corpora using the DAE loss (as opposed to just the author-specific corpus). 
The purpose of this baseline is to evaluate 
the content preservation capabilities of our setup.

\noindent\textbf{4. Supervised Stylized Rewriting}: \citeauthor{jhamtani2017shakespearizing} (\citeyear{jhamtani2017shakespearizing}) propose an LSTM-based encoder-decoder architecture for generating a ``Shakespearized'' text originally written in modern English, by leveraging parallel data. 
We compare this baseline only for generating Shakespearized text (using their data). We train the other three baselines and StyleLM by treating Shakespeare's corpus as the target author's corpus (without using the parallel nature of the data).

\begin{table*}[!h]
\centering
\scalebox{0.77}{
\begin{tabular}{| l | l | c | c | c | c |c | c | c | c |}
  \hline\multirow{2}{*}{\textbf{Data Source}} & \multirow{2}{*}{\textbf{Model}} & \multicolumn{5}{c|}{\textbf{Content Preservation} ($\uparrow$)} & \multicolumn{3}{c |}{\textbf{Stylistic Alignment} ($\downarrow$)}\\
 {} & {} & {BLEU} & {ROUGE-1} & {ROUGE-2} & {ROUGE-3} & {ROUGE-L} & {Lexical (MSE)} & {Syntactic (JSD)} & {Surface (MSE)}\\\hline
 \multirow{4}{*}{\textbf{Opinosis}} & GPT-2 & $18.3 \scriptstyle{\pm 2.3}$ & $0.51 \scriptstyle{\pm 0.06}$ & $0.29\scriptstyle{\pm 0.09}$ & $0.20\scriptstyle{\pm 0.06}$ & $0.36\scriptstyle{\pm 0.08}$ & $0.48\scriptstyle{\pm 0.06}$ & $0.27\scriptstyle{\pm 0.09}$ & $0.45\scriptstyle{\pm 0.01}$ \\
 
 {} & GPT-2 (FT) & $24.3 \scriptstyle{\pm 1.6}$  & $0.58 \scriptstyle{\pm 0.07}$ & $0.36\scriptstyle{\pm 0.08}$ & $0.27\scriptstyle{\pm 0.11}$ & $0.42\scriptstyle{\pm 0.09}$ & $0.32\scriptstyle{\pm 0.08}$ & $0.23\scriptstyle{\pm 0.02}$ & $0.40\scriptstyle{\pm 0.03}$\\
 
 {} & LM + DAE & $41.1 \scriptstyle{\pm 1.3}$ & $\mathbf{0.77} \scriptstyle{\pm 0.11}$ & $0.49\scriptstyle{\pm 0.05}$ & $0.39\scriptstyle{\pm 0.07}$ & $0.61\scriptstyle{\pm 0.08}$ & $0.33\scriptstyle{\pm 0.05}$ & $0.23\scriptstyle{\pm 0.01}$ & $0.38\scriptstyle{\pm 0.02}$\\
 
 {} & StyleLM & $\mathbf{43.4} \scriptstyle{\pm 1.7}$ & $0.73 \scriptstyle{\pm 0.13}$ & $\mathbf{0.53}\scriptstyle{\pm 0.06}$ & $\mathbf{0.41}\scriptstyle{\pm 0.08}$ & $\mathbf{0.68}\scriptstyle{\pm 0.07}$ & $\mathbf{0.29}\scriptstyle{\pm 0.04}$ & $\mathbf{0.19}\scriptstyle{\pm 0.01}$ & $\mathbf{0.31}\scriptstyle{\pm 0.04}$\\\hline

 \multirow{4}{*}{\textbf{Mark Twain}} & GPT-2 & $16.7 \scriptstyle{\pm 2.4}$ & $0.43 \scriptstyle{\pm 0.03}$ & $0.26\scriptstyle{\pm 0.07}$ & $0.16\scriptstyle{\pm 0.04}$ & $0.29\scriptstyle{\pm 0.09}$ & ${0.41}\scriptstyle{\pm 0.08}$ & $0.29\scriptstyle{\pm 0.03}$ & $0.42\scriptstyle{\pm 0.05}$ \\
 
 {} & GPT-2 (FT) & $22.9 \scriptstyle{\pm 1.6}$ & $0.49 \scriptstyle{\pm 0.06}$ & $0.38\scriptstyle{\pm 0.08}$ & $0.21\scriptstyle{\pm 0.06}$ & $0.37\scriptstyle{\pm 0.08}$ & ${0.35}\scriptstyle{\pm 0.07}$ & $0.25\scriptstyle{\pm 0.02}$ & $0.39\scriptstyle{\pm 0.06}$\\
 
 {} & LM + DAE & $31.7 \scriptstyle{\pm 1.5}$ & $\mathbf{0.68} \scriptstyle{\pm 0.14}$ & $0.44\scriptstyle{\pm 0.07}$ & $0.27\scriptstyle{\pm 0.07}$ & $0.45\scriptstyle{\pm 0.10}$ & $0.37\scriptstyle{\pm 0.03}$ & $0.24\scriptstyle{\pm 0.01}$ & $0.37\scriptstyle{\pm 0.03}$\\
 
 {} & StyleLM & $\mathbf{34.4} \scriptstyle{\pm 1.8}$ & $0.61 \scriptstyle{\pm 0.16}$ & $\mathbf{0.48}\scriptstyle{\pm 0.06}$ & $\mathbf{0.31}\scriptstyle{\pm 0.06}$ & $\mathbf{0.53}\scriptstyle{\pm 0.08}$ & $\mathbf{0.32}\scriptstyle{\pm 0.03}$ & $\mathbf{0.21}\scriptstyle{\pm 0.02}$ & $\mathbf{0.33}\scriptstyle{\pm 0.03}$\\\hline
 
 \multirow{4}{*}{\textbf{AI Wiki}} & GPT-2 & $12.6 \scriptstyle{\pm 2.1} $ & $0.37 \scriptstyle{\pm 0.04}$ & $0.19\scriptstyle{\pm 0.09}$ & $0.09\scriptstyle{\pm 0.05}$ & $0.25\scriptstyle{\pm 0.08}$ & $0.49\scriptstyle{\pm 0.07}$ & $0.31\scriptstyle{\pm 0.02}$ & $0.46\scriptstyle{\pm 0.05}$ \\
 
 {} & GPT-2 (FT) & $15.4 \scriptstyle{\pm 1.5}$ & $0.43 \scriptstyle{\pm 0.09}$ & $0.23\scriptstyle{\pm 0.06}$ & $0.13\scriptstyle{\pm 0.04}$ & $0.29\scriptstyle{\pm 0.07}$ & $0.40\scriptstyle{\pm 0.03}$ & $0.28\scriptstyle{\pm 0.03}$ & $0.42\scriptstyle{\pm 0.05}$\\
 
 {} & LM + DAE & $23.7 \scriptstyle{\pm 1.6}$ & $\mathbf{0.59} \scriptstyle{\pm 0.12}$ & $0.31\scriptstyle{\pm 0.08}$ & $0.18\scriptstyle{\pm 0.06}$ & $0.37\scriptstyle{\pm 0.09}$ & $0.41\scriptstyle{\pm 0.04}$ & $0.26\scriptstyle{\pm 0.02}$ & $0.41\scriptstyle{\pm 0.03}$\\
 
 {} & StyleLM & $\mathbf{26.7} \scriptstyle{\pm 1.9}$ & $0.54 \scriptstyle{\pm 0.13}$ & $\mathbf{0.34}\scriptstyle{\pm 0.08}$ & $\mathbf{0.23}\scriptstyle{\pm 0.08}$ & $\mathbf{0.46}\scriptstyle{\pm 0.09}$ & $\mathbf{0.34}\scriptstyle{\pm 0.02}$ & $\mathbf{0.22}\scriptstyle{\pm 0.01}$ & $\mathbf{0.36}\scriptstyle{\pm 0.04}$\\\hline
\end{tabular}}
\caption{\small{\textbf{Evaluating content preservation and stylistic alignment.}} \small{We evaluate the performance of \textit{StyleLM} against three baselines and on three test sets across multiple content preservation and stylistic alignment metrics. The reported numbers are mean and standard deviations ($\mu \pm \sigma$) across all the $10$ target authors. FT denotes author-specific fine-tuning; $\uparrow$ / $\downarrow$ indicates that higher / lower is better, respectively.}}
\label{tab:results}
\end{table*}

\begin{table*}[!h]
\centering 
\scalebox{0.73}{
\begin{tabular}{ | l | c | c | c | c |c | c | c | c |}\hline
  \multirow{2}{*}{\textbf{Model}} & \multicolumn{5}{c|}{\textbf{Content Preservation} ($\uparrow$)} & \multicolumn{3}{c |}{\textbf{Stylistic Alignment} ($\downarrow$)}\\
 {} & {BLEU} & {ROUGE-1} & {ROUGE-2} & {ROUGE-3} & {ROUGE-L} & {Lexical {(MSE)}} & {Syntactic {(JSD)}} & {Surface {(MSE)}}\\\hline
GPT-2 & $18.1$  & $0.43$ & $0.18$ & $0.11$ & $0.22$ & $0.41$ & $0.30$ & $0.43$\\
GPT-2 (FT) & $21.3$  & $0.48$ & $0.24$ & $0.16$ & $0.28$ & $0.36$ & $0.26$ & $0.39$\\
LM + DAE & $30.2$  & $0.55$ & $0.30$ & $0.19$ & $0.38$ & $0.32$ & $0.24$ & $0.36$\\
\citeauthor{jhamtani2017shakespearizing} (\citeyear{jhamtani2017shakespearizing}) & $31.3 $  & $\mathbf{0.57}$ & $\mathbf{0.33}$ & $0.23$ & $0.43$ & $0.29$ & $\mathbf{0.17}$ & $\mathbf{0.33}$\\
StyleLM & $\mathbf{33.8} $  & $0.53$ & $0.31$ & $\mathbf{0.25}$ & $\mathbf{0.44}$ & $\mathbf{0.28}$ & $0.21$ & ${0.34}$\\
\hline
\end{tabular}}
\caption{\small{\textbf{Comparison against supervised baseline. } Similar to Table \ref{tab:results}, we evaluate the performance of all the models against the approach of \cite{jhamtani2017shakespearizing} which relies on parallel data. For author-specific fine-tuning of \textit{StyleLM} and GPT-2 (FT), we use Shakespeare's corpus but without exploiting its parallel nature with modern English corpus.} }
\label{tab:results_PARALLEL}
\end{table*}

\subsection{Proposed Evaluation Methodology}
Following existing literature on style transfer and stylized text generation, we evaluate our proposed frameworks along two axes: \textit{content preservation} and \textit{stylistic alignment}.


\textbf{Content preservation} aims to measure the degree to which the generated stylized outputs have the same meaning as the corresponding input sentences. Following existing literature, we use the BLEU metric\footnote{BLEU score is measured with \textit{multi-bleu-detok.perl}} \cite{papineni2002bleu} and the ROUGE scores (ROUGE-1, ROUGE-2, ROUGE-3 and ROUGE-L \cite{rouge}). 

The core contribution of our evaluation framework is in the linguistic-motivation used to quantify the \textbf{stylistic alignment} of a generated piece of text with the target style we wish to achieve. While there have been several studies around formality and sentiment transfer on text, the same evaluation criteria does not apply to our setting because of two reasons: \textit{(a)} the classifier-based evaluation, which is facilitated by readily available classifiers for aspects like sentiment and formality, cannot be used to evaluate stylistic alignment with respect to an author's style, and \textit{(b)} author style is an amalgam of several linguistic aspects which are much more granular than the psycholinguistic concepts. To this end, taking motivation from \citeauthor{verma2019lexical} (\citeyear{verma2019lexical}), we formulate a multi-level evaluation scheme that identifies and quantifies stylistic expression at \textit{surface, lexical and syntactic} level. Once we quantify the stylistic expression, we use standard distance metrics to measure the stylistic alignment with target.


Linguists have identified style, especially in English language, to be expressed at three levels -- surface, lexical and syntactic \cite{strunk2007elements,dimarco1988stylistic,crystal2016investigating}. We first discuss the expression of stylistic elements as well their quantification. After quantifying the stylistic expressions at these levels, we discuss their incorporation into out evaluation framework.  

\textbf{\textit{Lexical elements}} of style are expressed at the \textit{word-level}. For instance, an authors choice words may be more subjective than objective (\textit{home} vs. \textit{residence}), or more formal than informal (\textit{palatable} vs. \textit{tasty}). For instance, we found that Rudyard Kipling, known for his classics of children's literature, had a higher tendency to use more concrete words (like, \textit{gongs}, \textit{rockets}, \textit{torch}, etc.) unlike Abraham Lincoln, who being a political writer, used more abstract words (like freedom, patriotism, etc.). Inspired from \citeauthor{brooke2013multi} (\citeyear{brooke2013multi}), we consider four different spectrums to take lexical-style into account: \textit{(i)} subjective-objective, \textit{(ii)} concrete-abstract, \textit{(iii)} literary-colloquial, and \textit{(iv)} formal-informal.

For quantifying these lexical elements, we use a list of seed words for each of the eight categories above, viz. subjective, objective, concrete, abstract, literary, colloquial, formal and informal \cite{brooke2013multi}. Following \citeauthor{brooke2013multi} (\citeyear{brooke2013multi}), we compute normalized  pointwise  mutual  information  index  (PMI)  to  obtain  a  raw  style score for each dimension,  by  leveraging  co-occurrences  of words in the large  corpus. The raw scores are  normalized to  obtain  style  vectors  for  every  word,  followed by a transformation of style vectors into k-Nearest Neighbor (kNN) graphs, where label propagation is applied. Since the eight original dimensions lie on the two extremes of four different spectrums, i.e., subjective-objective, concrete-abstract, literary-colloquial, and formal-informal, we compute $4$ averages across the entire author-specific corpus. The averages, in the range $[0, 1]$, denote the tendency of author using subjective, concrete, literary, or formal words, in contrast to using objective, abstract, colloquial, or informal words, as evidenced in their historical works\footnote{The final output is a $4$ dimensional vector with each of the elements, let's say $l_{sub} \in [0, 1]. $. The value of $l_{sub}$ will denote the tendency of the author to choose subjective words instead of their objective counterparts, which can be given by $1 - l_{sub}$}.

\textbf{\textit{Syntactic elements}} relate to the syntax of the sentence -- while some authors construct complex sentences, others construct simple sentences. For instance, as per the writings of Abraham Lincoln available in the Gutenberg corpus, a majority of his sentences can be categorized as compound-complex, while those of Rudyard Kipling's are mostly simple sentences (which are better suited to children). Taking inspiration from \citeauthor{feng2012characterizing} (\citeyear{feng2012characterizing}),  we categorize syntactic style into $5$ different categories -- (a) simple (b) compound (c) complex (d) complex-compound sentences, (e) others. For quantifying these stylistic elements, we compute the  fraction of sentences that are categorized into the $5$ categories by the algorithm proposed by \citeauthor{feng2012characterizing} (\citeyear{feng2012characterizing}). Since any given sentence will definitely lie in only one of the $5$ categories, the $5$ dimensional vector averaged across the sentences in a corpus can be thought of as probability distribution over the $5$ categories.

\textbf{\textit{Surface elements}} relate to statistical observations concerning aspects like the  average  number  of \textit{(i)} commas, \textit{(ii)} semicolons, \textit{(iii)} colons per sentence, \textit{(iv)} sentences in a paragraph, and \textit{(v)} number of words in a sentence. We quantify the surface-level elements into a $5$ dimensional vector.

Although the above enumerations of stylistic elements within a level, whether lexical, syntactic or surface, are not exhaustive, they are indicative of the stylistic expression at different levels. 
Computing the above statistics on an author-specific corpus gives an interpretable notion of the concerned author's writing style. Such a notion of style spans across multiple linguistic levels and has a considerable granularity. To this end, to quantify the stylistic alignment between generated text and the target text, we first compute these statistics for both the generated corpus and the target author's corpus. Then, we use standard distance metrics to obtain the extent of stylistic alignment at different linguistic levels. For lexical and surface-level alignment, we use mean squared error (MSE). Since syntactic style vector is a probability distribution over different syntactic categories, we use Jensen-Shannon divergence (otherwise known as symmetric KL divergence) to measure the alignment.

\begin{table*}[ht]
    \small
    \begin{tabular}{| c |  p{3.7cm} | p{3.5cm} | p{3.5cm} | p{3.4cm} |}\hline
        \textbf{Source} & {\textbf{Original Text}} & \textbf{NH's Style} & \textbf{CD's Style} & \textbf{GAH's Style}\\\hline
        \multirow{3}{*}{\textbf{Opinosis}} & {The staff was so polite and \textbf{catered} to our every need .} & {The staff was so polite and \textcolor{darkgreen}{kind} to our every need .} & {The staff was so polite and \textcolor{red}{obliged} to our every need .} & {The staff was so polite and \textcolor{blue}{ready to accept} our every need .}\\
        \cline{2-5}{} & {Front desk staff were \textbf{not super easy} to work with but ...} & {\textcolor{darkgreen}{Western desk , the} staff were \textcolor{darkgreen}{not abilities easy} to work with \textcolor{darkgreen}{,} but...} & {\textcolor{red}{front} desk \textcolor{red}{and} staff were \textcolor{red}{not extra easy} to work with \textcolor{red}{,} but...} & {The \textcolor{blue}{won} desk staff were \textcolor{blue}{not force easy} to work with \textcolor{blue}{,} but...}\\
        \hline
        \multirow{3}{*}{\textbf{Mark Twain}} & {I asked him \textbf{if he} learned to talk out of a book, and if I could \textbf{borrow it anywhere?}} & {I asked him \textcolor{darkgreen}{whether he had} learned to talk of a \textcolor{darkgreen}{dream ,} and if I could borrow \textcolor{darkgreen}{it.}} & {I asked him if he \textcolor{red}{had} learned to talk out of a book \textcolor{red}{;} and if I could borrow \textcolor{red}{it .}} & {I asked him if he learned to talk out of a \textcolor{blue}{man's mind} and if I could borrow \textcolor{blue}{it}}\\
        \cline{2-5}{} & {\textbf{Meanwhile}, if we understand each other now, I will go to work again.} & {\textcolor{darkgreen}{And} if we understand \textcolor{darkgreen}{each other's}, \textcolor{darkgreen}{I go to work.}} & {\textcolor{red}{And} if we understand each \textcolor{red}{other ,} I will go to \textcolor{red}{work.}} & {\textcolor{blue}{Then} if we understand \textcolor{blue}{each other's words} I will go to \textcolor{blue}{work.}}\\
        \hline
        \multirow{3}{*}{\textbf{AI Wiki}} & {If the \textbf{AI is programmed} for ``reinforcement learning", goals can be implicitly induced by \textbf{rewarding} some types of behavior or punishing others.} & {If the \textcolor{darkgreen}{human mind is bosoms for Heaven's sake }, he can be implicitly induced by rewarded some types of behavior or \textcolor{darkgreen}{punishment.}} & {If the \textcolor{red}{brain is learn} for men's object can be implicitly \textcolor{red}{induced by gratification} some \textcolor{red}{kind} of behaviour or punishment's punish's} & {If the \textcolor{blue}{round is turn for one's point} he can be implicitly induced by \textcolor{blue}{done some type of conduct} or punishing.}\\
        \hline
    \end{tabular}
    \caption{\textbf{Samples of stylized text generated by \textit{StyleLM}.} The target authors are Nathaniel Hawthorne (NH), Charles Dickens (CD) and George Alfred Henty (GAH). The source text has been taken from Opinosis, Mark Twain and AI Wiki, as indicated.}
    \label{tab:qualResults}
\end{table*}



\section{Results and Analysis}
\subsubsection{Qualitative Evaluation}Table \ref{tab:qualResults} presents samples of author-stylized text generated using StyleLM for some of the authors. Key highlights include the switch between `kind', `obliged' and `ready to accept' for the source word `catered'. The modification of the word `super' -- which is used in a colloquial sense, to `extra' without sacrificing the semantic meaning, demonstrates author-specific adaptation across different time frames. Similar observation can be made by noting the adaptation of `AI is programmed' to `brain is to learn' and `rewarding` to `gratification` on fine-tuning for Charles Dickens' writing style. Qualitative assessment of the generated samples depict the efficacy of our approach by illustrating alignment with the target author's style as well as significant content preservation.


\subsubsection{Quantitative Evaluation}
Our evaluation framework assesses the capability of our proposed \textit{StyleLM} model across both content preservation and stylistic alignment metrics. 

The results for stylized rewriting of the test corpus to the various author's style ($10$ in total) are presented in in Table \ref{tab:results}.  
All the fine-tuned \textit{StyleLM} models are tested on a test set that spans different domains -- \textit{(a)} Opinosis \cite{ganesan2010opinosis} which contains sentences extracted from user reviews on a variety of topics from Tripadvisor (hotels), Edmunds.com (cars) and Amazon.com (various electronics), \textit{(b)} text from Mark Twain's books, and \textit{(c)} a Wikipedia page on \textit{Artificial Intelligence}\footnote{We did not include any of these in the pre-training nor in the fine-tuning stage. As such, our model has never seen this data.}. To reiterate, the objective is to \textit{rewrite} the above test corpora into a style that reflects the style of target author we fine-tuned for. The averaged values for all 10 authors, as well as the standard deviation, across both content preservation as well as stylistic alignment metrics, are given in Table \ref{tab:results}.

It can be inferred from Table \ref{tab:results} that in terms of stylistic alignment, GPT-2 (FT), i.e., author fine-tuned GPT-2,  performs comparable to LM + DAE, i.e., denoising LM with no author-specific fine-tuning, across all the three datasets and on each of the three stylistic levels. However, the content preservation for LM + DAE is better than that of GPT-2 (FT). The vanilla GPT-2, however, shows the least impressive in terms of both content preservation as well stylistic alignment.  Specifically, the poor performance on content preservation can be attributed to the fact that GPT-2 and GPT-2 (FT) are both trained for generating continuations of input prompts and not for the task of stylistic \textit{rewriting}. It is nonetheless encouraging to see that fine-tuning the GPT-2 language model on author-specific corpus, i.e., GPT-2 (FT), increases the extent of  stylistic alignment with target author's style, establishing GPT-2 (FT) as a competitive baseline to compare stylistic alignment against. 

While LM + DAE, i.e., denoising LM without author-specific fine-tuning, shows good performance in terms of content preservation and stylistic alignment, our proposed approach, \textit{StyleLM}, shows considerable gains across all the metrics, against the LM + DAE. This observation confirms our hypothesis that the author-specific fine-tuning using DAE loss teaches the model to better learn the stylistic characteristics of the target author. Consistency of results across the diverse test sets shows a broader coverage in terms of applicability of the presented results. 

Interestingly, we notice that ROUGE-1 scores for the baseline LM + DAE (without author fine-tuning) are slightly higher than those for \textit{StyleLM}. A closer inspection of the generated samples from the two models reveals that this is because the stylized generations of the former are not as structurally coherent as those of the latter; i.e., while the predicted words are more accurate, they are not predicted in the correct order. This is further substantiated by the higher values for ROUGE-2, ROUGE-3 and ROUGE-L scores. 

\subsubsection{Comparison with Supervised Approach}
While \textit{StyleLM} performs better than the other unsupervised stylized generation models as shown in Table \ref{tab:results}, it is critical to determine its performance w.r.t. the supervised approach proposed by \citeauthor{jhamtani2017shakespearizing} (\citeyear{jhamtani2017shakespearizing}). We compare their LSTM-based encoder-decoder approach with GPT-2, GPT-2 (FT), LM + DAE and \textit{StyleLM} after fine-tuning them on Shakespeare's corpus. As it can be inferred from the results presented in Table \ref{tab:results_PARALLEL}, \textit{StyleLM} performs better than the supervised approach in terms of BLEU, ROUGE-3, ROUGE-L, and lexical stylistic alignment. The performance, as quantified by rest of the metrics, is comparable to that of \cite{jhamtani2017shakespearizing}. Given that \textit{StyleLM} was trained without leveraging the parallel nature of the data, 
the results are promising and demonstrate the abilities of our proposed model in generating author-stylized text while preserving the original content. 





\section{Conclusion \& Future Work}
In this work, we address the task of author-stylized rewriting by proposing a novel approach that leverages the generalization capabilities of language models. Building on the top of language models, we fine-tune on target author's corpus using denoising autoencoder loss to allow for stylistic adaptation in the process of reconstruction, without relying on parallel data. We also propose a new interpretable framework to evaluate stylistic alignment at multiple linguistic levels. We show that our proposed approach is able to capture the stylistic characteristics of target authors while rewriting the input text and performs not only better than other relevant and competitive baselines, but is also competent to an entirely supervised approach that relies on parallel data. 

The linguistic understanding of style, on which the proposed evaluation framework is based, can be used to guide the process of generating stylized text. The process of generation can be tuned to comply with attributes of style at different levels by penalizing or rewarding the (mis)alignment with these elemental attributes of style. Our plan is to explore this in further details, as part of future work.




\bibliographystyle{aaai}
\bibliography{biblio}
\end{document}